\documentclass[runningheads]{llncs}

\usepackage{eccv}

\usepackage{eccvabbrv}

\usepackage{graphicx}
\usepackage{booktabs}
\usepackage{nicefrac}
\usepackage{multirow}
\usepackage{algorithm} 
\usepackage{algpseudocode} 
\usepackage{relsize}
\usepackage{wrapfig}
\usepackage{microtype}

\usepackage[accsupp]{axessibility}  

\usepackage{hyperref}

\usepackage{orcidlink}

\newcommand{\algcom}[1]{{\color{gray} {\Comment{#1}}}}

\begin{document}

\title{LookupViT: Compressing visual information to a limited number of tokens} 

\titlerunning{LookupViT}

\author{Rajat Koner\inst{1,2}$^*$\orcidlink{0000-0003-3441-8192} \and
Gagan Jain\inst{1}$^*$\orcidlink{0009-0007-8394-9543} \and \\
Prateek Jain\inst{1} \and
Volker Tresp\inst{2} \and 
Sujoy Paul\inst{1}}

\authorrunning{Koner et al.}

\institute{Google DeepMind \and Ludwig Maximilian University of Munich}

\maketitle
\def\thefootnote{*}\footnotetext{denotes equal contribution}\def\thefootnote{\arabic{footnote}}

\begin{abstract}

    Vision Transformers (ViT) have emerged as the de-facto choice for numerous industry grade vision solutions. But their inference cost can be prohibitive for many settings, as they compute self-attention in each layer which suffers from quadratic computational complexity in the number of tokens. On the other hand, spatial information in images and spatio-temporal information in videos is usually sparse and redundant. In this work, we introduce LookupViT, that aims to exploit this information sparsity to reduce ViT inference cost. LookupViT provides a novel general purpose vision transformer block that operates by compressing information from higher resolution tokens to a fixed number of tokens. These few compressed tokens undergo meticulous processing, while the higher-resolution tokens are passed through computationally cheaper layers. Information sharing between these two token sets is enabled through a bidirectional cross-attention mechanism. The approach offers multiple advantages - (a) easy to implement on standard ML accelerators (GPUs/TPUs) via standard high-level operators, (b) applicable to standard ViT and its variants, thus generalizes to various tasks, (c) can handle different tokenization and attention approaches. LookupViT also offers flexibility for the compressed tokens, enabling performance-computation trade-offs in a single trained model. We show LookupViT's effectiveness on multiple domains - (a) for image-classification  (ImageNet-1K and ImageNet-21K), (b) video classification (Kinetics400 and Something-Something V2), (c) image captioning (COCO-Captions) with a frozen encoder. LookupViT provides $2\times$ reduction in FLOPs while upholding or improving accuracy across these domains. In addition, LookupViT also demonstrates out-of-the-box robustness and generalization on image classification (ImageNet-C,R,A,O), improving by up to $4\%$ over ViT.

\end{abstract}

\section{Introduction}
\label{sec:intro}

Images and videos, the cornerstones of modern visual communication, possess an inherent characteristic: their information content is often sparse and exhibits significant redundancy. However, Vision Transformers (ViTs) \cite{dosovitskiy2020image}, despite their dominance across multiple vision tasks, do not exploit this redundancy and attend to every token in a homogenized way.    
This leads to quadratic computational complexity with respect to image size, hindering its applicability in real-time situations. To bridge this gap, there is a pressing need to efficiently compress visual information into a smaller, more computationally manageable set of tokens. Such representations would unlock the potential of ViTs for resource-constrained scenarios while preserving their flexibility and performance advantages which led to their widespread adoption in the field of computer vision.
\begin{figure}[!t]
    \centering
    \begin{subfigure}[b]{0.58\textwidth}
        \includegraphics[height=2.85cm]{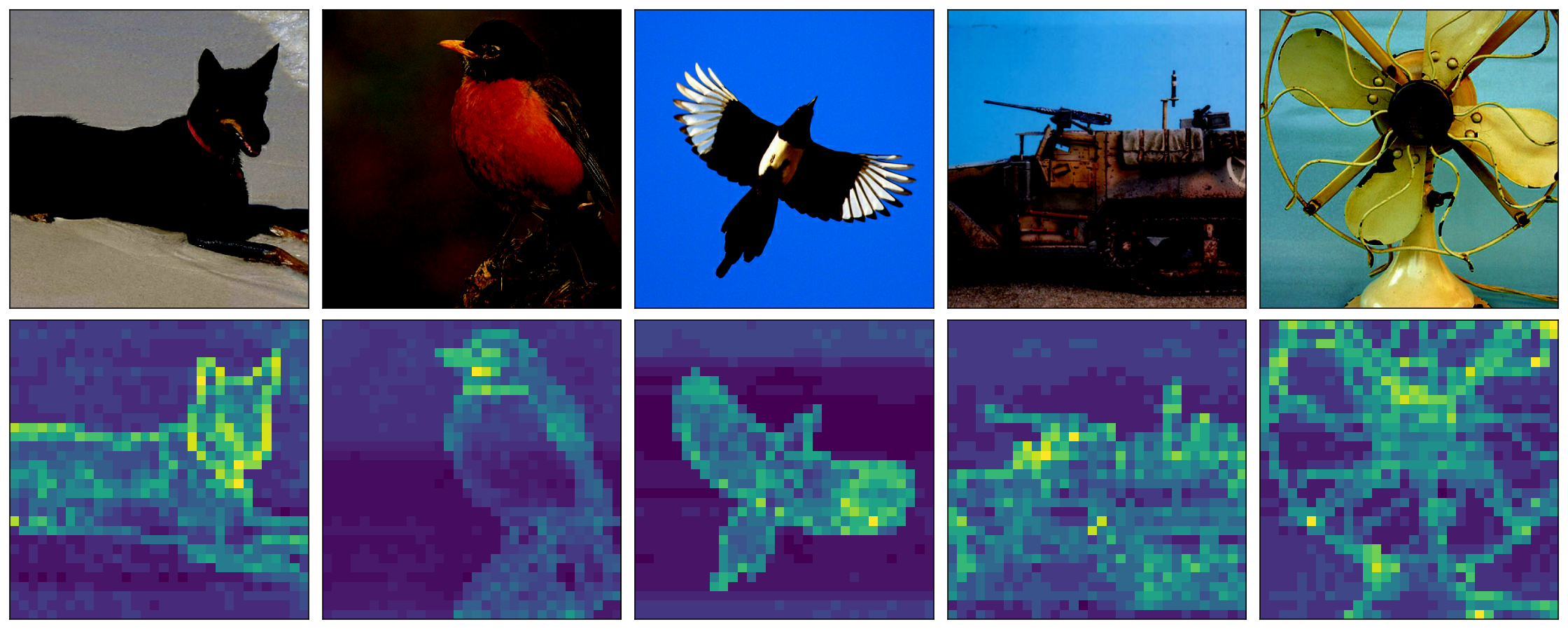}
        \caption{Cross-Attention Maps computed by LookupViT}
        \label{fig:attn_map}
    \end{subfigure}
    \hfill 
    \begin{subfigure}[b]{0.40\textwidth}
        \includegraphics[height=2.85cm]{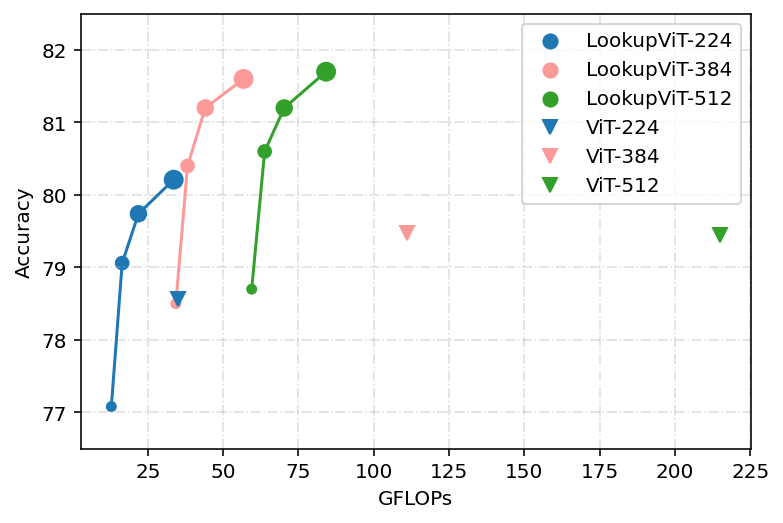}
        \caption{Performance with scaling image size}
        \label{fig:size_scaling}
    \end{subfigure}
    \caption{\small (a) Cross-attention maps between compressed and lookup tokens, emphasizing LookupViT's ability to extract relevant information from lookup tokens as needed for classification. (b) LookupViT vs ViT while scaling image resolution. The individual points per curve are for varied compressed tokens sizes ($3\times3, 5\times5, 7\times7, 10\times10$). LookupViT scales quite efficiently w.r.t ViT.}
    \label{fig:teaser}
\end{figure}


Several architectures aim to address the computational burden of ViTs by thoughtfully reducing the number of tokens. Token pruning methods retain a subset of tokens \cite{fayyaz2022adaptive,song2022cp,yu2023unified}, while token pooling techniques combine similar tokens for a more compact representation \cite{renggli2022learning,bolya2022token}. These mechanisms rely on heuristics derived from attention scores or feature similarities and might require additional task-specific adjustments. While these techniques offer valuable benefits, they may necessitate further fine-tuning based on the application. In contrast, we propose a novel LookupViT block to replace the vanilla ViT block, which intrinsically acts as a compression module. This design eliminates the need for post-processing or extensive fine-tuning. Furthermore, our method preserves the general structure of the ViT architecture, thus allowing further optimization and adaptations using existing approaches like token pruning or merging.

Compression modules like TokenLearner \cite{ryoo2021tokenlearner} and Perceiver \cite{jaegle2021perceiver} have also been explored in the literature. TokenLearner utilizes vanilla ViT blocks for a significant portion of the network depth, compressing a large number of tokens to a smaller set (e.g., 8 or 16) at later stages. This reliance on ViT blocks incurs a substantial computation  and heavily limits the the full utilization of compression module within the network. Perceiver, on the other hand, devises an asymmetric information flow directly from image pixels to a small set of latent representations iteratively throughout the network. Moreover, for these network architectures, it is non-trivial to extract multiple models with the same parameters, to exhibit a compute-performance trade-off between extracted models. LookupViT distinguishes itself by offering a scalable, computationally efficient block that can be seamlessly repeated like standard ViT blocks. Its bidirectional cross-attention mechanism facilitates a richer exchange of information between the compressed and original tokens, enhancing representational power. 

\begin{wrapfigure}{r}{0.5\textwidth}
    \centering
    \includegraphics[width=0.5\textwidth]{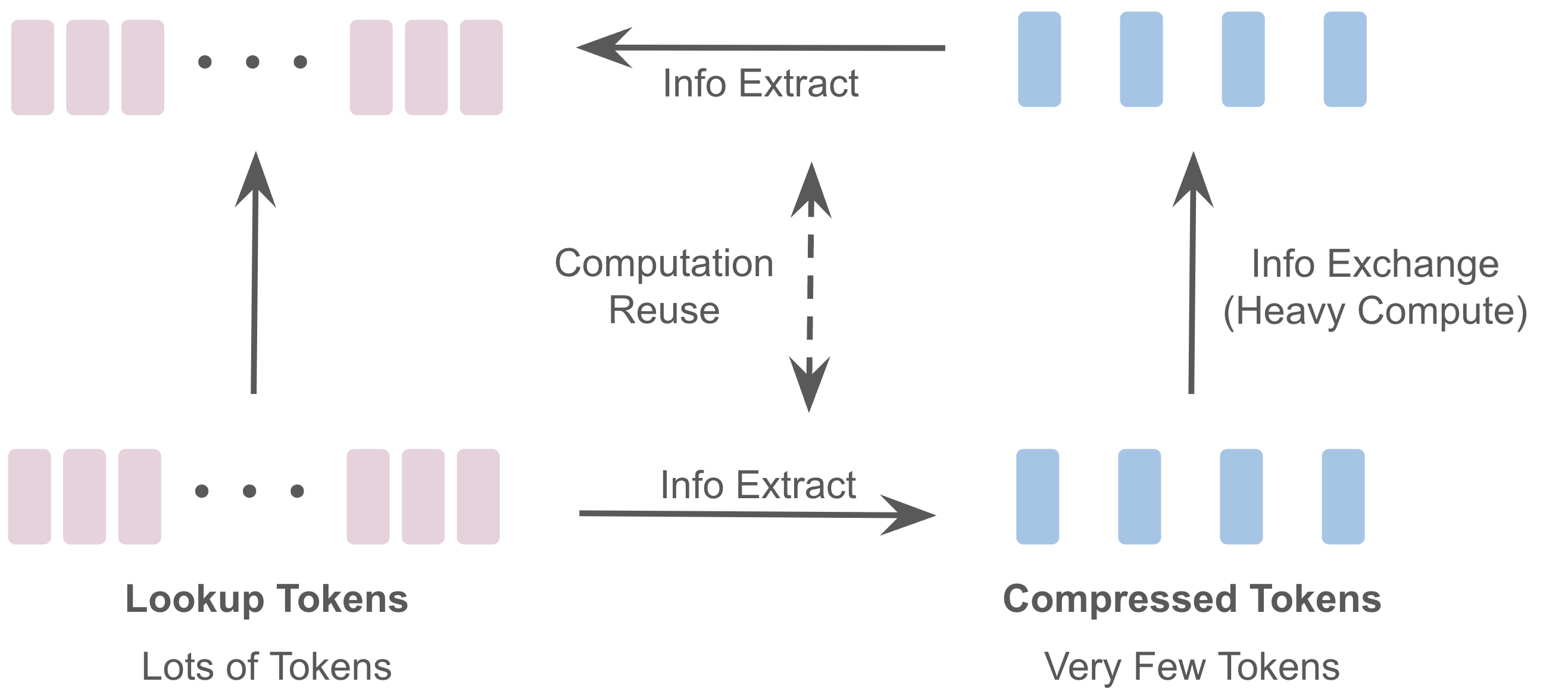}
    \caption{\small \textbf{Bidirectional information flow in LookupViT block.} LookupViT restricts the heavy computation to the compressed tokens, while extracting information from the lookup tokens. The lookup tokens then update themselves by reusing the information exchange computation.}
    \label{fig:flow}
\end{wrapfigure}

In this paper, we corroborate that for innately redundant  modalities like vision, condensing relevant spatial (and temporal) information from original tokens to a much smaller set can still sustain performance while significantly lowering the computational requirements, by maintaining an effective exchange of information between the two token sets. Figure \ref{fig:size_scaling} indicates LookupViT's ability to scale to large image sizes efficiently, by processing only relevant information, compared to vanilla ViT blocks, which scales quadratically in the number of original image tokens. We denote the smaller compressed set of tokens as \textit{compressed} tokens, which "look" at the larger original set of tokens, which we call \textit{lookup} tokens. The information exchange between these tokens happens in every LookupViT block in three key steps, as shown in Figure \ref{fig:flow} - (i) cross attention to transfer relevant information from the lookup tokens to the compressed tokens (shown in Figure \ref{fig:attn_map}), (ii) self-attention amongst the compressed tokens, and (iii) information transfer from the compressed tokens to the lookup tokens using shared attention weights, computed in the first step. While the compressed tokens communicate through self-attention, the lookup tokens communicate among themselves only via the compressed tokens. This technique avoids the quadratic scaling, while ensuring that the lookup latent representations get richer along the layers.

LookupViT's intrinsic design naturally supports flexibility in terms of token compression and variable image or token size. By adjusting the down-sampling ratio between compressed and lookup tokens, the cost-performance trade-off can be tailored to match specific application requirements. This multi-resolution nature allows for extraction of compute-efficient high-performing models during inference, with the same parameter space. To validate LookupViT's efficacy, we show results on multiple benchmarks like image and video classification, and image captioning. Notably, due to the information bottleneck, LookupViT also shows out-of-the-box robustness to image corruptions. The key contributions of this work are -
\begin{itemize}
    \item \textbf{Efficient Token Compression}: LookupViT introduces a novel Multi-Head Bidirectional Cross-attention (MHBC) module that enables effective information flow with significant computational savings.
    \item \textbf{Generalized Framework}: LookupViT offers a flexible framework applicable to visual modalities. It also offers compute-performance trade-offs via multi-resolution ability of compressed tokens, with identical model parameters.
    \item \textbf{Enhanced Performance}: LookupViT generalizes well to applications on image/video modalities, and boasts out-of-the box robustness to corruptions.
\end{itemize}

\section{Related Works}
Since the introduction of the Vision Transformer (ViT), a multitude of works have endeavored to improve its efficiency and scalability. 
{\flushleft \textbf{Multi-scale and Hierarchical Features:}} Early studies such as \cite{liu2021swin, fan2021multiscale, wang2021pyramid} utilized non-overlapping patches with multi-scale or hierarchical features, achieving notable success in both image and video domains \cite{arnab2021vivit}. Concurrently, \cite{ryali2023hiera} proposed hierarchical designs for efficient training and inference across these modalities. These approaches pushed accuracy boundaries, but often at the expense of added architectural complexity. For instance, MViTv2 \cite{li2022mvitv2} decomposes relative position embedding and residual pooling, while CSWin \cite{dong2022cswin} integrates cross-shaped windows within a hierarchical framework. This creates a trade-off between enhanced accuracy and the potential loss of ViT's inherent simplicity and scalability. LookupViT's compressed and lookup tokens has some parallels with the convolution-based OctConv’s \cite{chen2019drop} low and high frequency features. However LookupViT restricts heavy processing to compressed tokens, and enjoys scalability of Transformers. 
{\flushleft \textbf{Token Merging and Sampling:}} Another prominent research direction involves token merging and pruning. \cite{bolya2022token, fayyaz2022adaptive, song2022cp, yu2023unified} aim to reduce redundant tokens through merging, sampling, or pruning. For example, \cite{bolya2022token} uses similarity to groups and merge tokens, while \cite{fayyaz2022adaptive} employs adaptable token sampling. While valuable, these techniques often introduce heuristics and generally function as post-processing steps.  Furthermore, they can face challenges when extending to modalities beyond images, such as videos or multi-modal data. In contrast, LookupViT emphasizes intrinsic compression through its core architecture, replacing the ViT block.  Importantly, LookupViT remains harmonious with the potential application of token merging or sampling for further optimization.

{\flushleft \textbf{Token compression:}} Instead of merging tokens, \cite{renggli2022learning} learns a smaller number of M patches from the original N patches in ViT using a learnable weight matrix. Similarly, TokenLearner \cite{ryoo2021tokenlearner} compressed all ViT tokens into a smaller set of 8-16 tokens and performing self-attention within this reduced set, but after a certain number of vanilla ViT layers. Perceiver \cite{jaegle2021perceiver} proposes learning a small set of tokens directly from the pixel space using iterative unidirectional cross-attention. These two methods are most closely related to our work. However, TokenLearner's compression achieves optimal performance only when processing at least  $50-75\%$ of the network with ViT blocks, leading to no reduction in computation for a significant number of layers. In contrast, LookupViT can be trained entirely with lookup blocks, reducing computational complexity without compromising performance. Furthermore, unlike Perceiver \cite{jaegle2021perceiver}, which uses unidirectional pixel-level cross-attention, LookupViT operates on tokens with bi-directional cross-attention to update both compressed and lookup tokens.

{\flushleft \textbf{Flexible patch and resolution:}} Recent works like FlexiViT \cite{beyer2023flexivit} address the fixed patch size limitation by training with multiple patch sizes, enabling ViT to scale across different patch sizes and image resolutions. Na-ViT \cite{dehghani2024patch} explores sequence packing to train images with arbitrary resolution and aspect ratio, allowing inference on any resolution image. Analogous to these works, we show that LookupViT can also be trained with varying compression ratios to obtain multiple models during inference with the same parameter space. 



\section{LookupViT Methodology}
\label{sec:method}
In this section, we discuss the LookupViT framework in detail, starting with a high-level architectural discussion, and then focusing on specific design choices. We also discuss its applicability to downstream tasks and Multi-Resolution flexibility. We conclude this section with an analysis of the improved computational complexity.

\begin{figure}[t!]
    \centering
    \includegraphics[width=0.9\textwidth]{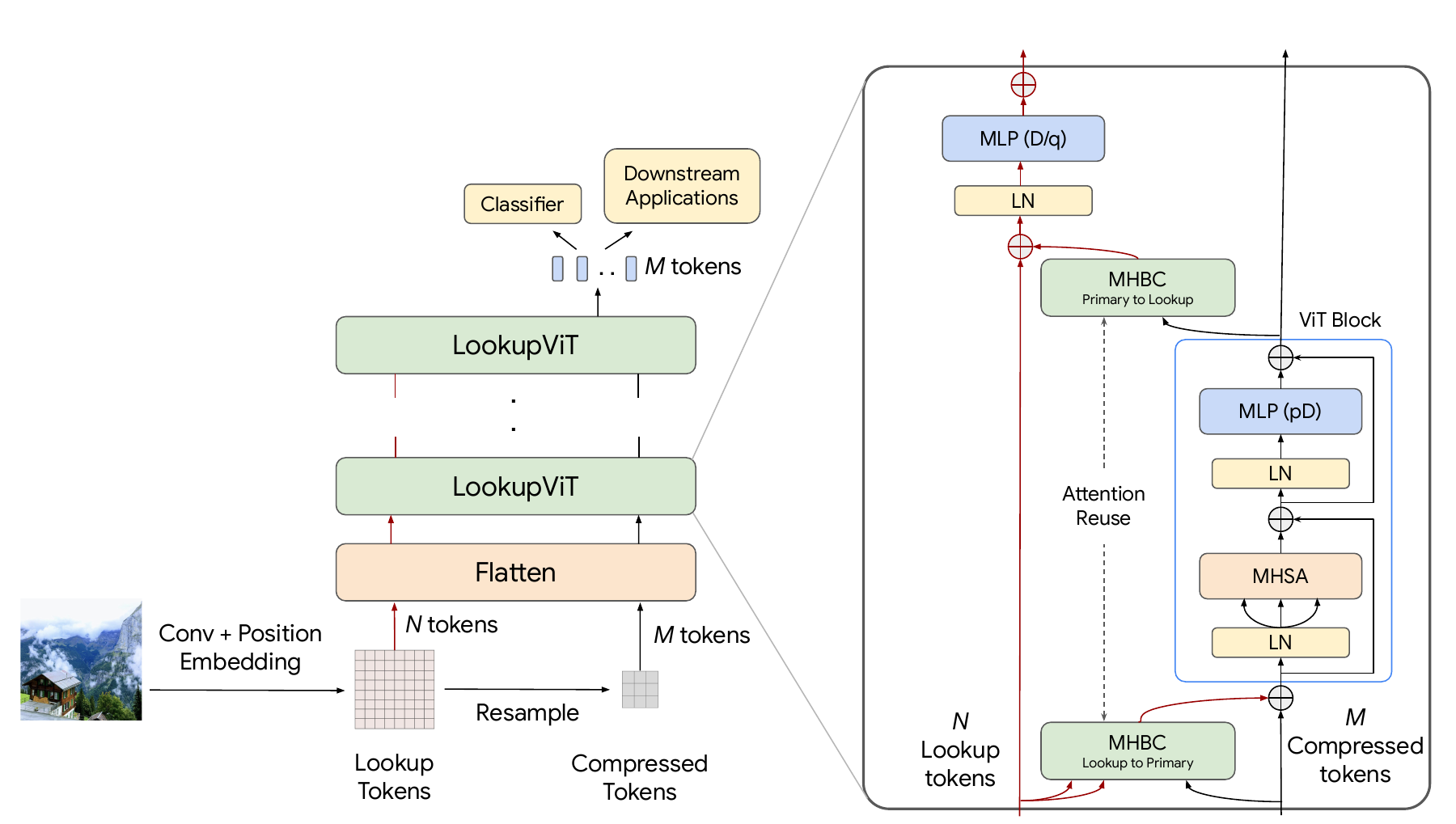}
    \caption{\small \textbf{LookupViT Architecture:} The LookupViT block  is stacked multiple times similar to vanilla ViT. Each LookupViT block has two parallel computation streams for the two different types of tokens. Heavy computation happens on a fixed smaller number of compressed tokens, while light computation happens on the much higher number of lookup tokens. There is an asynchronous information exchange between the two token sets using the Multi-Head Bi-Directional Cross Attention (MHBC) block.}
    \label{fig:lookupvit_framework}
\end{figure}

\subsection{Overall Architecture}
\label{subsec:overall}
An overview of the LookupViT architecture is presented in Figure \ref{fig:lookupvit_framework}. Similar to the ViT architecture, it comprises of a stack of LookupViT blocks.  First, an input RGB image (or video) is divided into non-overlapping patches. These patches are then passed through a convolutional layer to generate feature embeddings. Positional embeddings are then added to construct the input 
tokens – a process identical to the standard ViT architecture \cite{dosovitskiy2020image}. Unlike vanilla ViT, the core idea here is - \textit{to compress visual information into a smaller number of tokens, focusing heavy computation exclusively on those tokens.}

A fixed number of tokens $M$ $(\ll N)$, which we name as the \textit{compressed} tokens are sampled from the input tokens, using bilinear interpolation. Computationally intensive processing is performed on the compressed tokens, analogous to a standard ViT block, while exchanging information with the original tokens through asynchronous Multi-Head Bidirectional Cross-Attention (MHBC). The process unfolds as follows - (1) \textbf{Information Gathering:} Compressed tokens use cross-attention to ``look" at the original tokens (termed \textit{lookup} tokens) and gather relevant information.
(2) \textbf{Representation Refinement:} Compressed tokens exchange information amongst themselves, updating their representations. (3) \textbf{Global Context Infusion:} The lookup tokens utilize the processed, information-rich compressed tokens, to update their own representations, reusing the attention weights calculated during Information Gathering for efficiency.


During this entire process, the lookup tokens are forced to gather  information only by interacting with the compressed tokens, thus reducing computational complexity.  Additionally, the lookup tokens pass through a MLP block with a smaller projection dimension ($D/q$) compared to the vanilla model projection ($pD$), which is applied on the compressed tokens, where $D$ represents the transformer embedding dimension ($(p,q) = (4,2)$). This optimization further reduces computations. The LookupViT block's ability to achieve performance comparable to the baseline, despite this substantial MLP bottleneck, demonstrates the effectiveness of the information exchange between compressed and lookup tokens.



\subsection{Input Tokenization} 
\label{subsec:input_tok}

The construction of lookup token embeddings similar to standard ViT \cite{dosovitskiy2020image} tokenization strategy. Given an input image $\boldsymbol{\mathrm{X}} \in \mathbb{R}^{h \times w \times c}$, it is passed through a convolutional layer to obtain lookup features $\boldsymbol{\mathrm{F}}_l \in \mathbb{R}^{h_l \times w_l \times D}$. A learnable lookup positional embedding $\boldsymbol{\mathrm{F}}_{l,pos} \in \mathbb{R}^{h_l \times w_l \times D}$ is added to this feature map. These tokens are then significantly downsampled to a fixed shape -- $(h_p, w_p)$, which constitute the compressed tokens. This can be summarized as below - 
{\small\begin{align}
    \boldsymbol{\mathrm{F}}_p &\leftarrow \boldsymbol{\mathcal{T}}\big(\boldsymbol{\mathrm{F}}_l, (h_p, w_p)\big) & \boldsymbol{\mathrm{F}}&_{p, pos} \leftarrow \boldsymbol{\mathcal{T}}\big(\boldsymbol{\mathrm{F}}_{l, pos}, (h_p, w_p)\big)\\
    \boldsymbol{\mathrm{F}}_{p} &\leftarrow \boldsymbol{\mathrm{F}}_{p} + \boldsymbol{\mathrm{F}}_{p, pos} & \boldsymbol{\mathrm{F}}&_{l} \leftarrow \boldsymbol{\mathrm{F}}_{l} + \boldsymbol{\mathrm{F}}_{l, pos}
\end{align}}
The operator $\boldsymbol{\mathcal{T}}(\mathbf{x},s)$ bilinearly resizes $\mathbf{x}$ to shape $s$. The lookup and compressed token grids have sizes $(h_l, w_l)$ and $(h_p, w_p)$, and $D$ is the embedding dimension. These feature maps $\boldsymbol{\mathrm{F}}_{p}$ and $\boldsymbol{\mathrm{F}}_{l}$ are then spatially flattened to $\boldsymbol{\mathrm{z}}^0_p$ and $\boldsymbol{\mathrm{z}}^0_l$:
{\small\begin{align}
& \boldsymbol{\mathrm{z}}^0_p = [\boldsymbol{\mathrm{F}}_{p(0,0)}, \dots, \boldsymbol{\mathrm{F}}_{p(h_p-1,w_p-1)}] & \boldsymbol{\mathrm{z}}^0_p &\in \mathbb{R}^{h_p.w_p \times D}\\
& \boldsymbol{\mathrm{z}}^0_l = [\boldsymbol{\mathrm{F}}_{l(0,0)}, \dots, \boldsymbol{\mathrm{F}}_{l(h_l-1,w_l-1)}] & \boldsymbol{\mathrm{z}}^0_l &\in \mathbb{R}^{h_l.w_l \times D}
\end{align}}
These flattened feature maps $\boldsymbol{\mathrm{z}}^0_p$ and $\boldsymbol{\mathrm{z}}^0_l$ (compressed and lookup tokens respectively) are passed as input to the LookupViT block, which efficiently refines these representations through information exchange, as explained in Section \ref{subsec:lvit_block}. The resize ratio $C = h_l.w_l/h_p.w_p$ is a flexibility parameter, representing the degree of information compression. This enables us to flexibly train the model with varying resize ratio, thus allowing compute-aware model extraction with a specific $C$. A smaller value for $C$ indicates more number of compressed tokens and thus better representation power. In fact, $C=1$ would represent the vanilla ViT with certain extra computations due to the cross-attention. We denote the number of lookup and compressed tokens by $N = h_l.w_l$ and $M = h_p.w_p$ respectively. This form of tokenization can readily extend to videos, where a third dimension representing time would be introduced. The compression ratio would then become $C = h_l.w_l.t_l/h_p.w_p.t_p$, where $t_{.}$ denotes the temporal dimension. 

\subsection{LookupViT Block}

\label{subsec:lvit_block}
The $k^{th}$ LookupViT block consumes the compressed tokens $\boldsymbol{\mathrm{z}}^{k-1}_p$  and lookup tokens $\boldsymbol{\mathrm{z}}^{k-1}_l$ from its previous block, facilitates information exchange between the two token sets, and passes the updated representations to the next block. The novel architectural design here is the asynchronous Multi-Head Bidirectional Cross-attention (MHBC).
Intuitively, in the first layer, the lookup tokens maintain a richer image representation than the compressed tokens. However, after multiple passes through the LookupViT block, the compressed tokens accumulate relevant compressed image information, thus making them suitable for downstream tasks. This happens through iterative communication between the lookup and compressed tokens in every LookupViT block (Algorithm \ref{alg:lookupvitblock}). This can be summarized into three key steps -

{\flushleft \textbf{Information Gathering:}} In this step, there is a unidirectional information flow from the lookup to the compressed tokens through MHBC$_{l\rightarrow p}$. The compressed tokens are used as query ($\boldsymbol{\mathrm{Q}}$) and lookup tokens as key-value ($\boldsymbol{\mathrm{K}},\boldsymbol{\mathrm{V}}$). Algorithm \ref{alg:MHBC1} presents this part of the proposed MHBC module. Additionally, we store the attention weights $\mathcal{A}$ computed in this step to be re-used while sharing information in the reverse direction.


{\flushleft \textbf{Representation Refinement:}}
After the information extraction step, the compressed tokens go through a vanilla ViT block (self-attention followed by MLP), as illustrated in Algorithm \ref{alg:vitblock}. The MLP dimension upscaling factor $p$ is kept equal to 4, as in vanilla ViT. But this computation happens on the smaller compressed token set. This step allows internal information sharing between compressed tokens to update their representation. 

{\flushleft \textbf{Global Context Infusion:}}
The information gathering along with the ViT based processing enriches the compressed token features, as they contain a  compressed global representation of the image. While the lookup tokens do not directly share information amongst themselves, they are notified about the global information through a reverse direction information exchange, from compressed to lookup tokens, as depicted in Algorithm \ref{alg:MHBC2} (MHBC$_{l\rightarrow p}$). Rather than recomputing the attention matrix, we reuse the attention matrix previously saved  in MHBC$_{p\rightarrow l}$. This relation further imposes implicit similarity constraints between the two feature maps, and enhances information exchange. Finally, to refine the lookup features, we apply a low dimensional MLP block, with a dimension ($D/q)$,  $pq$ times smaller than the vanilla ViT MLP dimension (we set $(p, q) = (4, 2)$ in all our experiments). This enriches the lookup tokens for information extraction by the compressed tokens in the next LookupViT block. 

{\raggedleft
 \begin{minipage}[][][b]{0.46\linewidth}
\begin{algorithm}[H]
	\caption{\small MHBC$_{l\rightarrow p}$} 
    \label{alg:MHBC1}
    \scriptsize
    \textbf{In:} $\boldsymbol{\mathrm{z}}_p \in \mathbb{R}^{M \times D}$; $\boldsymbol{\mathrm{z}}_l \in \mathbb{R}^{N \times D}$
    \begin{algorithmic}[1]
	    \State $Q \leftarrow$ LN($w_Q \boldsymbol{\mathrm{z}}_p$) 
	    \State $K \leftarrow$ LN($ w_K \boldsymbol{\mathrm{z}}_l$) 
	    \State $V \leftarrow w_V \boldsymbol{\mathrm{z}}_l$ 
	    \State $\mathcal{A}\leftarrow$ softmax($QK^T$) \algcom{$\mathcal{A} \in \mathbb{R}^{M\times N}$}
	    \State $\boldsymbol{\mathrm{z}}_p\leftarrow\boldsymbol{\mathrm{z}}_p + \mathcal{A}V$ 
	\end{algorithmic} 
	\textbf{Return} $\boldsymbol{\mathrm{z}}_p, {\mathcal{A}}$
\end{algorithm}
\end{minipage}
\hfill
\begin{minipage}{0.48\linewidth}
\begin{algorithm}[H]
	\caption{\small MHBC$_{p\rightarrow l}$}  
    \label{alg:MHBC2}
    \scriptsize
    \textbf{In:} $\boldsymbol{\mathrm{z}}_l \in \mathbb{R}^{N \times D}$; $\boldsymbol{\mathrm{z}}_p \in \mathbb{R}^{M \times D}$; $\mathcal{A} \in \mathbb{R}^{M\times N}$\\
	\begin{algorithmic}[1]
	    \State $V \leftarrow$ LN($w_V \boldsymbol{\mathrm{z}}_p$) 
	    \State \algcom{Layer Norm on Values}
	    \State $\boldsymbol{\mathrm{z}}_l\leftarrow\boldsymbol{\mathrm{z}}_l + \mathcal{A}^TV$ 
	    \State \algcom{Reuse pre-computed weights}
	\end{algorithmic} 
	\textbf{Return} $\boldsymbol{\mathrm{z}}_l$
\end{algorithm}
\end{minipage}
}
{\raggedleft\noindent\begin{minipage}{0.48\textwidth}
\begin{algorithm}[H]
	\caption{\small ViTBlock} 
	\scriptsize
	\label{alg:vitblock}
    \textbf{In:} $\boldsymbol{\mathrm{z}}_p \in \mathbb{R}^{M \times D}$; $p \in \mathbb{N}$\\
	\begin{algorithmic}[1]
	    \State $\boldsymbol{\mathrm{z}}_p \leftarrow \boldsymbol{\mathrm{z}}_p + $ MHSA\big(LN($\boldsymbol{\mathrm{z}}_p$)\big)
	    \State \algcom{Multi-Head Self-Attention}
	    \State $\boldsymbol{\mathrm{z}}_p\leftarrow\boldsymbol{\mathrm{z}}_p + $ MLP\big(LN($\boldsymbol{\mathrm{z}}_p$); $pD$\big)
	\end{algorithmic} 
	\textbf{Return} $\boldsymbol{\mathrm{z}}_p$
\end{algorithm}
\end{minipage}
\hfill
\begin{minipage}{0.48\linewidth}
\begin{algorithm}[H]
	\caption{\small LookupViTBlock} 
	\scriptsize
    \label{alg:lookupvitblock}
    \textbf{In:} $\boldsymbol{\mathrm{z}}_p \in \mathbb{R}^{M \times D}$; $\boldsymbol{\mathrm{z}}_l \in \mathbb{R}^{N \times D}$; $p, q \in \mathbb{N}$
	\begin{algorithmic}[1]
	    \State $\boldsymbol{\mathrm{z}}_p, {\mathcal{A}} \leftarrow$ MHBC$_{l\rightarrow p}$\big(LN($\boldsymbol{\mathrm{z}}_p$), LN($\boldsymbol{\mathrm{z}}_l$)\big)
	    \State $\boldsymbol{\mathrm{z}}_p\leftarrow$ViTBlock\big($\boldsymbol{\mathrm{z}}_p$, $p$\big)
	    \State $\boldsymbol{\mathrm{z}}_l\leftarrow\boldsymbol{\mathrm{z}}_l + $ MHBC$_{p\rightarrow l}$\big(LN($\boldsymbol{\mathrm{z}}_p$), LN($\boldsymbol{\mathrm{z}}_l$), ${\mathcal{A}}$\big) 
	    \State $\boldsymbol{\mathrm{z}}_l\leftarrow\boldsymbol{\mathrm{z}}_l + $ MLP\big(LN($\boldsymbol{\mathrm{z}}_l$);$D/q$\big) 
	\end{algorithmic} 
	\textbf{Return} $\boldsymbol{\mathrm{z}}_p$, $\boldsymbol{\mathrm{z}}_l$
\end{algorithm}
\end{minipage}
}

\subsubsection{Multi-Resolution Tokens:} The compressed tokens are constructed by simply resizing the lookup tokens in a non-learnable fashion. Hence, it is possible to share the same parameter space and lookup tokens while having multiple compressed token resolutions. To do this, we choose the compressed token size uniformly at random during training, seeking inspiration from FlexiViT \cite{beyer2023flexivit}. Once trained in this fashion, we can extract multiple high performing models having different computational requirements from a single trained model. This flexibility makes our method utilizable in a variety of settings, depending on resource availability.

\subsection{Training and Token Utilization for Downstream Applications}
In LookupViT, we maintain two sets of tokens throughout the network - $N$ lookup tokens and $M$ compressed tokens. For classification, we can apply the classifier to either or both token sets. Empirically, we've found that enforcing classification loss on both heads yields the best performance. We use global average pooling on the respective token sets, followed by two separate classifiers. The joint loss function is then optimized with equal weights. 

Although the training loss is applied independently to both token sets, we find that during inference, the classifier on the compressed tokens is sufficient. However, adding the classifier output from the lookup tokens does improve performance marginally. Since there is no added computational cost for classification, we average the outputs of both compressed and lookup heads with equal weights. For downstream applications beyond classification (e.g., vision-language model tasks like captioning), a decoder is used on the LookupViT encoder. In such cases, using a limited compressed token set computationally benefits the cross-attention block. Hence, we experiment using only the compressed tokens.



\subsection{Computational Complexity}

Let $\mathcal{C}_x$ denote the computation of a procedure $x$. Then, given the feature dimension $D$, number of lookup tokens $N$, number of compressed tokens $M (<<N)$, MLP upscaling factor $p=4$ (on compressed tokens) and downscaling factor $q=2$ (on lookup tokens), the computational complexity of the vanilla ViT and LookupViT blocks can be represented as follows (neglecting smaller terms).

{\small\begin{align}
{\mathlarger{\mathcal{C}}}_{\mathrm{ViT}} &= 2N^2D + 12ND^2\\
{\mathlarger{\mathcal{C}}}_{\mathrm{LookupViT}} &= \left(3NM + 2M^2\right)D + \left(4N + 15M\right)D^2
\end{align}
}
Notice that we get rid of the quadratic dependence on the number of lookup tokens $N$ and reduce the attention and linear projection computations individually. Since the number of compressed tokens $M (<< N)$ stay constant at a user-specified value, the attention reduction factor grows quickly, enabling scalability for usage at higher resolutions. Typically, for an image resolution of 384, we use $N=576$ and $M=25$, which shows superior performance than the vanilla model, while simultaneously reducing FLOPs by a factor greater than $3$.

\section{Results}
\label{sec:results}

{\flushleft \textbf{Implementation Details:}}
As ViTs are prone to overfit more as compared to CNNs, they either need pre-training on large datasets like JFT \cite{sun2017revisiting} or augmentation based training frameworks like DeIT \cite{touvron2021training} or AugReg \cite{steiner2021train}. Due to the ease of implementation and adaptability to other tasks that we pursue in this work, we build our implementation on top of \cite{steiner2021train}. We implement LookupViT in JAX \cite{jax2018github} within the Big Vision repository \cite{big_vision}. We adopt the exact training settings as in \cite{steiner2021train} (like learning rate, training epochs, etc) without performing any parameter sweeps. We also train TokenLearner \cite{ryoo2021tokenlearner}, another state-of-the-art token compression technique, on the same repository for fair comparison, with $16$ tokens for all experiments. TokenLearner$_{1/2}$ denotes their compression module is applied half-way through the network, which the authors recommend. 


{\flushleft \textbf{Image classification:}}
We evaluate LookupViT on image classification while -- (a) training from scratch on ImageNet-1k \cite{deng2009imagenet}, and (b) finetuning on ImageNet-1k from a ImageNet-21k pre-trained model. The popular benchmark ImageNet-1k has $1.28$ million training images and $50,000$ validation images across $1,000$ categories. ImageNet-21k has 12.8 million images across $21,000$ categories. For all experiments, we train and report performance on the validation set with an image size of $224\times 224$, unless specified otherwise. We experiment with two model sizes $B$ and $L$, with model parameters as defined in ViT \cite{dosovitskiy2020image}. 

We present the results on image classification in Table \ref{table:imagenet}. LookupViT is flexible enough to offer multiple models with the same parameter, by varying the compression factor $C$, the ratio between the number of lookup and compressed tokens. Results indicate that  training from scratch with the B/16 model on ImageNet-1k, LookupViT$_{5\times 5}$ performs better than ViT with $2.12\times$ lesser FLOPs. LookupViT$_{10\times 10}$ outperforms ViT by $1.6\%$ while still being computationally cheaper. It also offers similar gains compared to TokenLearner and Perceiver. On image size of $384$, we can see in Figure \ref{fig:size_scaling}, LookupViT offers more than $3\times$ computational gains compared to ViT. The figure also shows that LookupViT computationally scales quite efficiently w.r.t. ViT when the image size is increased. 




\begin{table}[t!]
\centering
\caption{Comparison with state-of-the-art methods on ImageNet-1k and ImageNet-21K pre-training followed by ImageNet-1K finetuning.}
\scriptsize
\setlength{\tabcolsep}{6pt}
\label{table:imagenet}
\begin{tabular}{clccc}
\toprule
Variant & Method                      & ImageNet-1K & ImageNet-21K & GFLOPs \\
\midrule
 \parbox[t]{6mm}{\multirow{7}{*}{\small B/16}} 
& ViT \cite{dosovitskiy2020image} & 78.6 & 83.7  & 35.1 \\ 
&TokenLearner$_{1/2}$ \cite{ryoo2021tokenlearner}     & 75.7 & 82.0 &  19.1  \\ 
&TokenLearner$_{3/4}$ \cite{ryoo2021tokenlearner}     & 77.5 & 82.9 &  27.1  \\ 
&Perceiver \cite{jaegle2021perceiver}           & 78.0 & - &  707  \\ 
&LookupViT$_{3\times3}$         & 77.1  & 77.9 & 12.9   \\ 
&LookupViT$_{5\times5}$        & 79.1 & 81.6 & 16.5  \\ 
&LookupViT$_{7\times7}$       & 79.7  & 83.0 & 21.9   \\ 
&LookupViT$_{10\times10}$      & 80.2 & 83.9 & 33.6 \\
\midrule
 \parbox[t]{6mm}{\multirow{6}{*}{\small L/16}} 
 & ViT \cite{dosovitskiy2020image}        & 75.7 & 84.0 & 123.5  \\ 
&TokenLearner$_{1/2}$ \cite{ryoo2021tokenlearner}     & 78.5 & 84.2 &  66.5  \\ 
&TokenLearner$_{3/4}$ \cite{ryoo2021tokenlearner}     & 77.6 & 84.6 &  94.6  \\ 
&LookupViT$_{3\times3}$      & 77.2 & 78.4  &  46.2\\ 
&LookupViT$_{5\times5}$        & 78.7 & 81.9  & 58.8  \\ 
&LookupViT$_{7\times7}$         & 79.1 & 83.3 & 77.7  \\ 
&LookupViT$_{10\times10}$       & 79.2 & 84.1 & 118.5  \\ 
\bottomrule
\end{tabular}
\end{table}

The performance of LookupViT on the large model is even better when trained from scratch. Even with $3\times 3$ compressed tokens, LookupViT performs much better than ViT, requiring $2.67\times$ lower FLOPs. It is interesting to note that we did observe instabilitites while training large models for both ViT and TokenLearner, whereas we do not observe such instabilities in LookupViT. When using ImageNet-21k pretrained models for ImageNet-1k finetuning, we achieve higher accuracy than the ViT models using our $10\times10$ models, while still maintaining lesser computational requirements.

{\flushleft \textbf{Analyzing the robustness of LookupViT:}}
Vision models often exhibit surprising vulnerability to image corruptions and adversarial perturbations. While the ViT architecture is more robust than CNNs in general \cite{bai2021transformers}, we explore out-of-the box robustness of LookupViT to image corruptions and adversarial settings, without including any additional robustness losses, augmentations or training strategies.  We evaluate on ImageNet-A,C,O,R \cite{hendrycks2021many, hendrycks2019robustness, hendrycks2021nae}. (see Appendix A1)

\begin{figure}[b!]
    \centering
    \begin{subfigure}[b]{0.4\textwidth}
        \includegraphics[width=\textwidth]{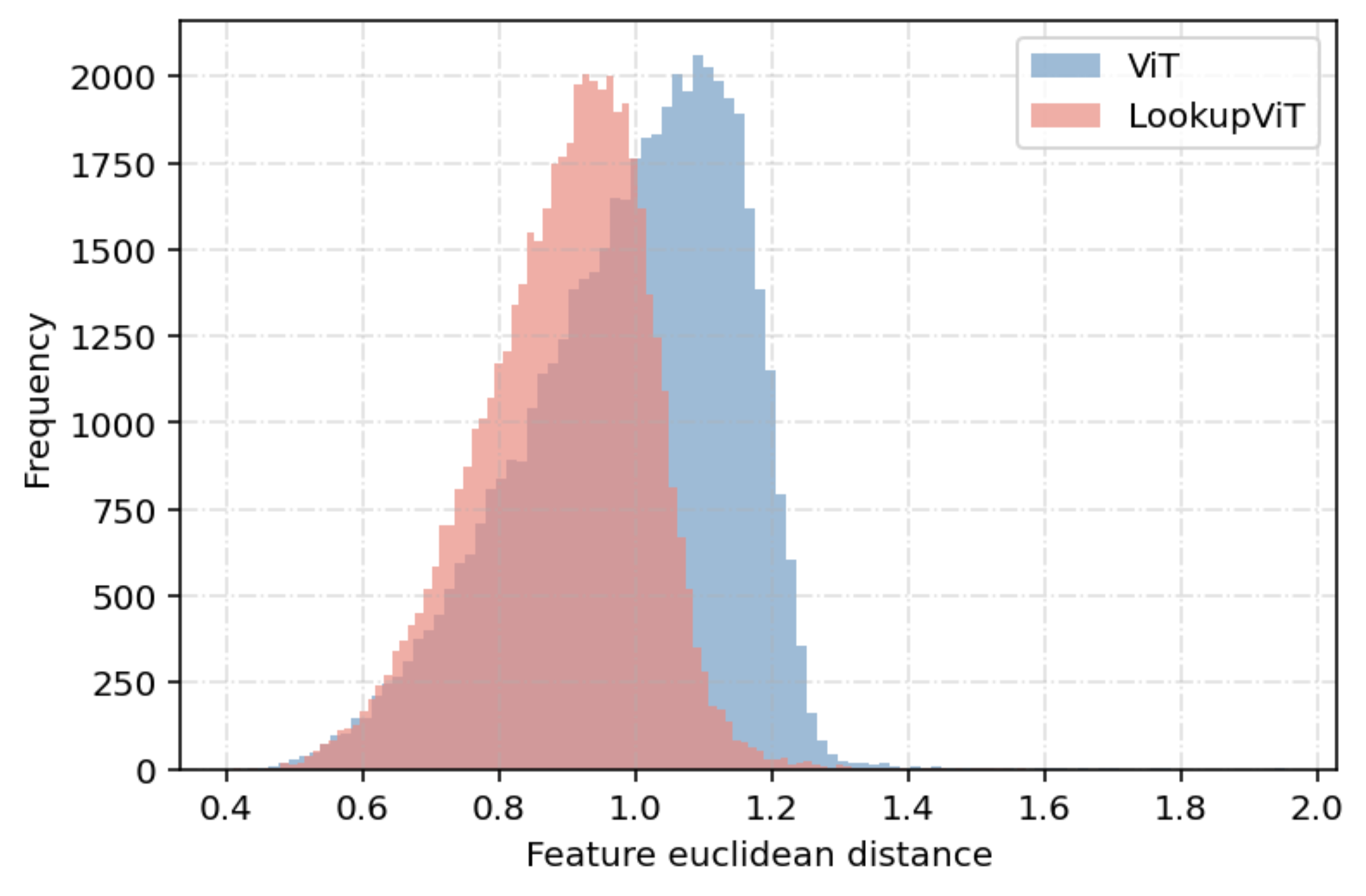}
        \caption{}
        \label{fig:avg_dist_density}
    \end{subfigure}
    \begin{subfigure}[b]{0.4\textwidth}
        \includegraphics[width=\textwidth]{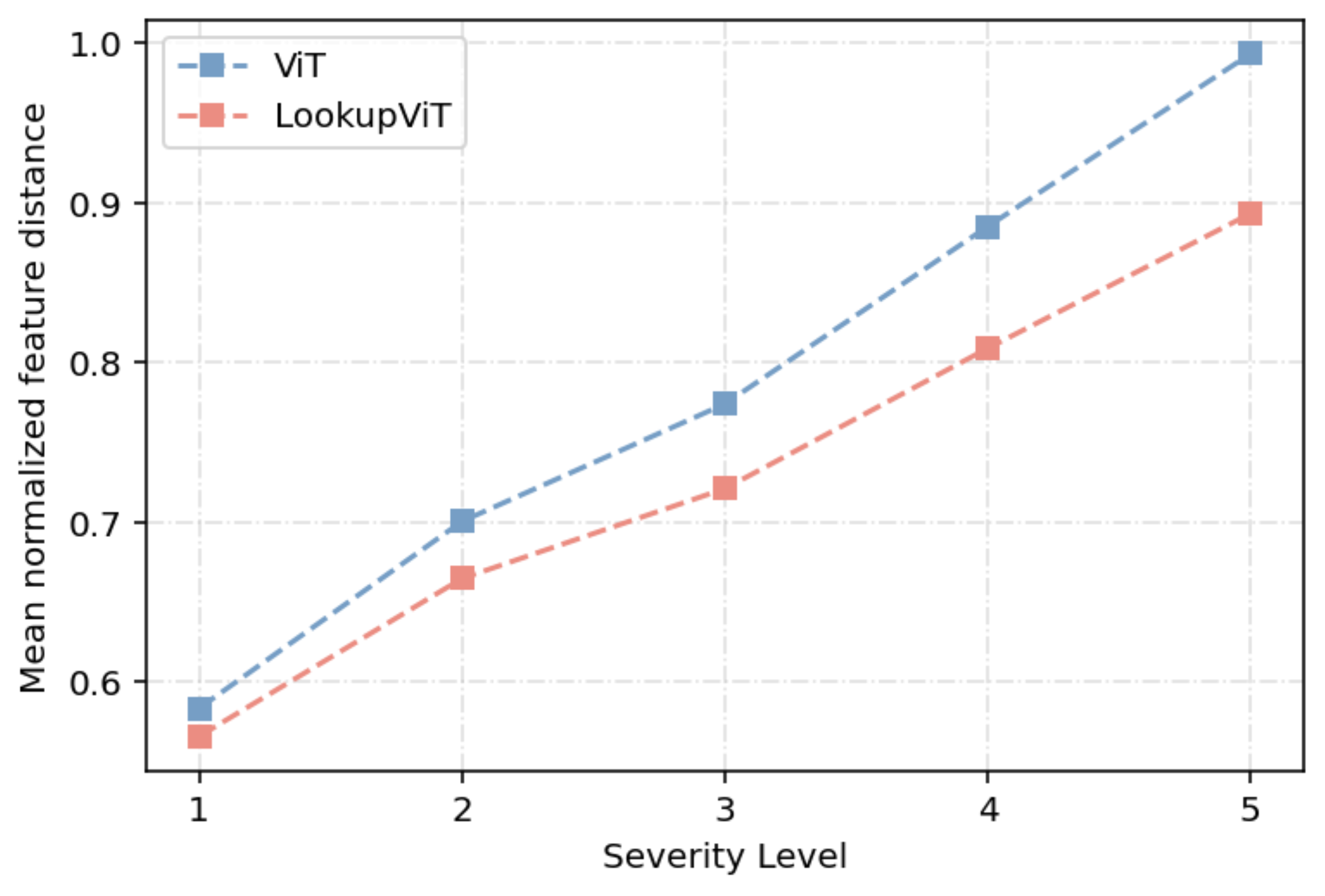}
        \caption{}
        \label{fig:feat_dist_sev}
    \end{subfigure}
    \vspace{-10pt}
    \caption{(a) Density of normalized feature distance for severity=5 over all corruptions. (b) Mean normalized feature distance over all corruptions for different severity.}
    \label{fig:robustness}
\end{figure}

As shown in Table \ref{table:corruptions}, LookupViT performs better than ViT across the board, showing robust performance on natural, unmodified images (ImageNet-A), resilience to image corruptions (ImageNet-C) and artistic renditions (ImageNet-R), and strong generalization to out-of-distribution samples (ImageNet-O). These unanimously suggests that LookupViT's mechanism of extracting only useful information inherently improves its ability to handle noisy or distorted inputs.

We further investigate LookupViT's robustness to perturbations by comparing the image-wise normalized feature deviation due to corruptions in Figure \ref{fig:robustness}, on the ImageNet-C dataset. It also shows that the margin of improvement of LookupViT over ViT goes up as we increase the corruption severity. This shows the model's robust behaviour beyond its better discriminatory ability.

\begin{table*}[t!]
\small
\centering
\caption{Robustness on ImageNet-A, C, O, R \cite{hendrycks2021many, hendrycks2019robustness, hendrycks2021nae} datasets.}
\label{table:corruptions}
\scriptsize
\setlength{\tabcolsep}{5pt}
\begin{tabular}{c|l|cccc}
\toprule
Variant & Method &  ImageNet-C & ImageNet-O & ImageNet-R & ImageNet-A\\
\midrule
\parbox[t]{6mm}{\multirow{4}{*}{\small B/16}} 
&ViT \cite{dosovitskiy2020image} & 56.4 & 24.6 & 25.8 & 6.7 \\ 
&TokenLearner$_{3/4}$ \cite{ryoo2021tokenlearner}  & 55.7 & 24.5 & 23.1 &  5.7\\ 
&LookupViT$_{5\times5}$        & 56.4 & 24.0 & 25.9 & 5.4 \\ 
&LookupViT$_{7\times7}$      & 58.4  & 24.9 & 28.0 & 7.6\\ 
&LookupViT$_{10\times10}$      & \textbf{59.9} &  \textbf{25.7} & \textbf{29.1} & \textbf{8.1}\\ 
\midrule
\parbox[t]{6mm}{\multirow{4}{*}{\small L/16}} 
&ViT \cite{dosovitskiy2020image}  & 55.3 & 22.4 & 19.4 & 3.5\\
&TokenLearner$_{3/4}$ \cite{ryoo2021tokenlearner} & 55.6  & 20.3 & 24.0 & 3.7 \\ 
&LookupViT$_{5\times5}$        & 58.4  & 25.7 & 26.5 & 7.0 \\ 
&LookupViT$_{7\times7}$       & 59.7  & \textbf{26.7} & 28.0 & 8.3\\ 
&LookupViT$_{10\times10}$     & \textbf{60.4}  & 26.5 & \textbf{28.1} & \textbf{8.4} \\ 

\bottomrule
\end{tabular}
\end{table*}

{\flushleft \textbf{Using pre-trained LookupViT for Captioning:}} In Table \ref{table:captions}, we assess the capability of LookupViT as a pre-trained model to judge its transferability for other tasks. We investigate its performance on image captioning using Locked Image Tuning (LiT) style training \cite{zhai2022lit}. Following LiT, we freeze the parameters of the pretrained LookupViT image encoder, which is pre-trained on ImageNet-21k. We then introduce a simple text decoder, initialized randomly, and train it to generate captions corresponding to the image representations produced by LookupViT. We perform this experiment on the COCO Captions \cite{chen2015microsoftcoco} dataset. 

LookupViT with $7\times 7$ compressed tokens exhibits similar performance to ViT, even without finetuning, thus offering significant reduction in FLOPs. This highlights the quality of visual representations learned by LookupViT and its potential as a versatile backbone for various vision-and-language tasks.

\begin{table}[bt!]
\centering
\caption{Image captioning on COCO-Captions \cite{chen2015microsoftcoco} using LiT decoder style training \cite{zhai2022lit} with frozen encoder. (LookupViT: LViT, TokenLearner: TL)}
\label{table:captions}
\scriptsize
\setlength{\tabcolsep}{3pt}
\begin{tabular}{l  c  c  c  c  c c}
\toprule
& ViT \cite{dosovitskiy2020image} & TL$_{1/2}$ \cite{ryoo2021tokenlearner} & LViT$_{3\times3}$ & LViT$_{5\times5}$  & LViT$_{7\times7}$  & LViT$_{10\times10}$  \\
\midrule
 Cider & 112.6 & 110.3 & 104.3 & 108.4 & 111.3 & 111.7 \\
 FLOPS &  59.5 & 38.2 & 31.9 & 35.9 & 42.1 & 55.3 \\
\bottomrule
\end{tabular}
\end{table}

{\flushleft \textbf{Video classification:}}
LookupViT can be easily extended to videos. We modify the ViViT \cite{arnab2021vivit} spatio-temporal B/16 encoder to construct LookupViViT. As in ViViT, the initial Conv3D layer, with kernel size $16\times16\times2$ operates on a video of $224\times224\times3\times32$. The resultant 3D tokens serve as lookup tokens, which are bilinearly downsampled to obtain the spatio-temporal compressed tokens. After flattening the tokens, the rest follows exactly as in LookupViT. 

We carry out experiments on the Kinetics400 \cite{kay2017kinetics} and SomethingSomethingV2 (SSv2) \cite{goyal2017something} datasets. Kinetics has 240k videos of 10 second duration each. Being a dynamic YouTube based dataset, it often  incurs data loss due to deletion, so we could only train on a subset of videos which ViViT was trained on at the time of its paper publication, leading to lower baselines. SSv2 has 220k videos of 2-6 second duration each. On SSv2, ViViT \cite{arnab2021vivit} does not report numbers using the Spatio-Temporal B/16 model, but we follow the training recipe as mentioned for their L/16 Factorized Encoder \cite{arnab2021vivit}, and initialize our LookupViViT as well as ViViT models from their corresponding Kinetics400 finetuned models.
\begin{figure}[b!]
    \centering
    \begin{subfigure}[b]{0.46\textwidth}
        \includegraphics[width=0.9\textwidth]{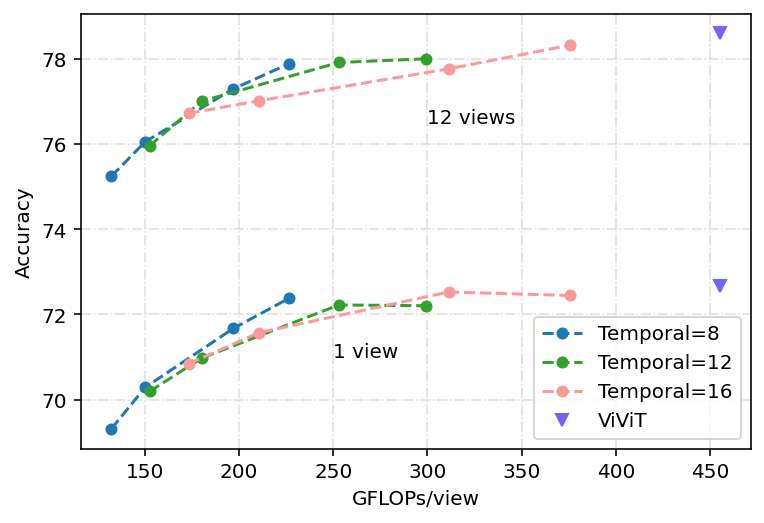}
        \caption{LViViT with different compressed tokens}
        \label{fig:video}
    \end{subfigure}
    \hfill 
    \begin{subfigure}[b]{0.46\textwidth}
        \includegraphics[width=0.9\textwidth]{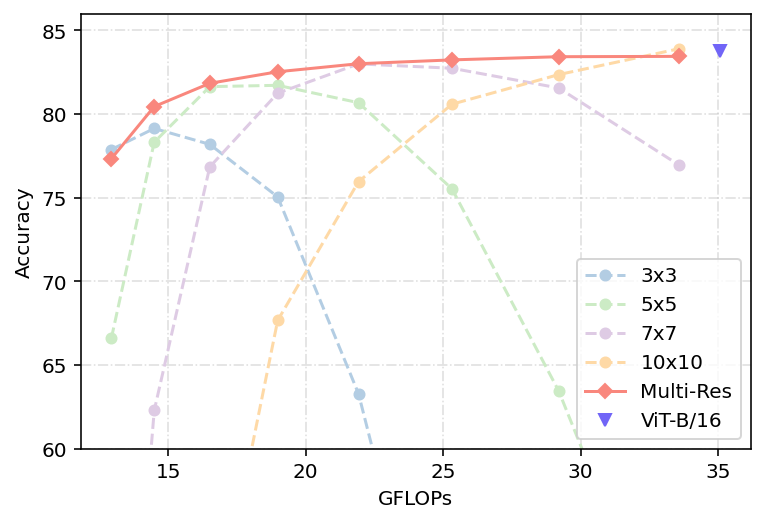}
        \caption{Multi-resolution compressed tokens}
        \label{fig:multires}
    \end{subfigure}
    \caption{(a) Video classification on K400 with different spatio-temporal compressed tokens for LookupViViT (LViViT). Color denotes the number of temporal token and points on each curve are with increasing number of spatial tokens. (b) Training a single model (``Multi-Res") which can handle different number of compressed tokens, offering compute-performance trade-off with the same parameter space. The other models are trained individually but evaluated at all resolutions.}
    \label{fig:combined}
\end{figure}

The results on video classification are presented in Table \ref{table:video} for both single and multi-crop evaluation following ViViT \cite{arnab2021vivit}. Moreover, we report results for various number of spatial and temporal compressed tokens and plot them in Figure \ref{fig:video} for Kinetics400. LookupViViT models with half the FLOPs as ViViT show competitive results on Kinetics400. We also found the trends in accuracy to be increasing with the number of spatial and/or temporal tokens, as expected.

Interestingly, as in Table \ref{table:video}, LookupViViT performs significantly better than ViViT on SSv2. We observe a $5\%-6\%$ improvement in accuracy with half the FLOPs. This further bolsters LookupViT's robustness claim, as in SSv2 backgrounds and objects are similar across classes, thus needing recognition of fine-grained motion \cite{arnab2021vivit}, which our model does better than ViViT. (See Appendix A2, A5) 

\begin{table*}[t!]
\centering
\caption{Comparison of LookupViT-B/16 based ViViT \cite{arnab2021vivit} model with state-of-the-art methods on Kinetics400 \cite{kay2017kinetics} and SomethingSomethingV2 \cite{goyal2017something}.}
\label{table:video}
\scriptsize
\setlength{\tabcolsep}{8pt}
\resizebox{0.9\textwidth}{!}{
\begin{tabular}{l cc | cc | c}
\toprule
\multirow{2}{*}{Method} & \multicolumn{2}{c}{Kinetics400} & \multicolumn{2}{c}{SomethingSomethingV2} & \multirow{2}{*}{GFLOPs/view} \\
                        & 1 view & 12 views & 1 view & 12 views  \\ 
\midrule
ViViT \cite{dosovitskiy2020image}           & 72.6  & 78.6  & 51.2 & 52.3 & 455.1  \\ 
LookupViViT$_{9\times9\times8}$          & 72.4 & 77.9  & 56.3 & 58.3 & 226.5\\ 
LookupViViT$_{8\times8\times16}$       & 72.5 & 77.8  & 57.0 & 59.2 & 311.2\\ 
LookupViViT$_{9\times9\times16}$    & 72.5 & 78.3   & 57.1 & 59.6 & 375.8 \\ 
\bottomrule
\end{tabular}
}
\end{table*}

{\flushleft \textbf{Multi-resolution LookupViT:}}
We empirically demonstrate the effectiveness of LookupViT's multi-resolution tokenization. By varying the downsampling ratio, we can control the trade-off between computational cost and representation capacity while keeping the parameter size constant. Inspired by \cite{beyer2023flexivit}, during training, we randomly choose the compressed token resolution, ranging from $3\times3$ to $10\times10$  for every batch of data. We call this model ``Multi-Res".  The number of lookup tokens is always kept fixed at $14\times 14$ for an image resolution of $224 \times 224$. To highlight its efficacy, we also train individual models with the designated number of compressed tokens, while evaluating at all compressed resolutions. 
\begin{table*}[bt!]
  \caption{\textbf{Dissecting the network:} Component-wise performance impact}
  \label{table:ablations}
  \scriptsize
\setlength{\tabcolsep}{8pt}
  \centering
    \begin{tabular}{lccc}
    \toprule
    Models &  GFLOPS & Accuracy \\
    \midrule
    LookupViT-B/16 &  16.5 & 79.1  \\
    $ \quad - $ No Lookup Tokens  & 4.5 & 69.2 \\
    $ \quad - $ No MHBC$_{p\rightarrow l}$ & 11.0 & 69.7 \\
    $ \quad - $ No Loss on Lookup Tokens & 16.5 & 78.2  \\
    $ \quad - $ No Loss on Primary Tokens & 16.5 & 78.4 \\
    $ \quad $ Random Primary Tokens & 16.5 & 77.9\\
    \bottomrule
    \end{tabular}
\end{table*}

We present results in Figure \ref{fig:multires}, where all models are pretrained on ImageNet-21k and finetuned on ImageNet-1k. For Multi-Res, both steps are carried out with a multi-resolution training technique. As we can see, the performance of the Multi-Res model is remarkably close to that of the individual models at the points for which they are trained. The performance of the individual models do not hold up when evaluated at other resolutions. This finding highlights the adaptability of LookupViT and its potential to streamline model selection by offering performance-computation trade-offs within a single trained architecture.

\section{Ablations}

 The ablations performed in this section use a B/16 model with a compressed token size $5 \times 5$ trained from scratch on ImageNet-1k. This model, with all the components in place, reaches a top-1 classification accuracy of 79.1. We discuss the component-wise importance of LookupViT model  in Table \ref{table:ablations}.

\textbf{No Lookup Tokens:} We consider constructing the compressed tokens by aggresively downsampling the image features through convolution, while having no information support through the higher resolution lookup tokens, i.e. no MHBC$_{l\rightarrow p}$  or MHBC$_{p\rightarrow l}$. The compressed tokens go through only the vanilla ViT. This leads to much lower performance, indicating that ViT, by itself, doesn't work well with very limited tokens without additional information exchange.

\textbf{No MHBC}$_{p\rightarrow l}$\textbf{:} From the previous step, we now add MHBC$_{l\rightarrow p}$, which facilitates information transfer from lookup to compressed tokens, while still not updating the lookup tokens. This leads to a slight increase in performance as compared to the previous setup, as the lookup tokens are not updated at all with global information. However, this step involves construction of compressed tokens using a parameter-free resize rather than a convolutional downsampling, which enables use of same model across different compressed token sizes.

\textbf{No Lookup/Compressed Loss:} Next, we add MHBC$_{p \rightarrow l}$, a source of information exchange from the compressed to the lookup tokens, with loss computation still only on the compressed tokens. This leads to a further $8.5\%$ increase in accuracy, thus justifying the need for the bidirectional MHBC. We also consider the case where we add loss on the lookup logits only, but not on the compressed, and this leads to an equivalent performance, indicating the near equal capability of compressed and lookup tokens.

\textbf{Random Compressed Tokens:} We also experiment with random learnable compressed tokens in the first layer, instead of resizing from lookup tokens. We observe a $\sim$1\% performance drop, thus showing the effectiveness of parameter-free resize operation for constructing compressed tokens.

\section{Conclusions}
In this work, we present a novel LookupViT architecture, which efficiently compresses sparse and redundant visual information to fewer tokens. By efficiently combining lower and higher resolution tokens with bidirectional cross-attention, LookupViT achieves a significant reduction in FLOPs while upholding performance of ViT. Its effectiveness is demonstrated on diverse vision tasks, like image and video classification, image captioning, as well as it generalizability and robustness to visual corruptions.

Future work includes extending our model to dense prediction tasks like object detection and semantic segmentation, as well as scaling to larger model sizes.

\clearpage  

%
%
\bibliographystyle{splncs04}
\bibliography{main}

\appendix
\section{Appendix}

In this section, we present detailed arguments indicating the robustness of our method, supported by additional results and visualisations.

\subsection{Robustness on ImageNet family of datasets}
\label{app:robustness}

The ImageNet family of datasets provides a comprehensive suite for evaluating the robustness of vision models. ImageNet-A assesses performance on real-world, unmodified images that are typically misclassified by models, gauging their ability to handle naturally occurring challenges. ImageNet-C introduces common image corruptions like blur and noise, measuring resilience to various degradations. ImageNet-R applies artistic styles to the original images, testing a model's ability to generalize across diverse visual renditions. ImageNet-O presents out-of-distribution samples from classes not found in the standard ImageNet-1k dataset, evaluating a model's robustness to unfamiliar objects and scenes. Together, these datasets offer a multi-faceted assessment of a vision model's performance, spanning natural challenges, degradations, artistic variations, and out-of-distribution generalization.

We further analyse results on ImageNet-C in a greater detail here. ImageNet-C consists of 15 corruption types applied across five severity levels. In Table \ref{table:severity}, we compare LookupViT's performance under these corruptions with a vanilla ViT model and TokenLearner, all trained on the standard ImageNet-1k dataset. We report the accuracy for all severities, along with their corresponding averages and mean Corruption Error (mCE) as introduced in \cite{hendrycks2019benchmarking}.

Table \ref{table:severity} indicates the superior performance of LookupViT with lower computational requirements in the presence of adverse perturbations, depicting superior robust behaviour than ViT and TokenLearner \cite{ryoo2021tokenlearner}.

\begin{table*}[bhtp!]
\small
\centering
\caption{Performance comparison on the corrupted ImageNet-C dataset \cite{hendrycks2019robustness}.}
\label{table:severity}
\scriptsize
\setlength{\tabcolsep}{4.5pt}
\begin{tabular}{c|l|cccccc|c}
\toprule
{Variant} & {Method} & sev1 & sev2 & sev3 & sev4 & sev5 & Avg. $\uparrow$ & mCE $\downarrow$\\
\midrule
\parbox[t]{6mm}{\multirow{4}{*}{\small B/16}} 
&ViT \cite{dosovitskiy2020image} & 70.3 & 64.1 & 59.0  &  49.9 & 38.9 & 56.4 & 55.3  \\ 
&TokenLearner$_{3/4}$ \cite{ryoo2021tokenlearner}  & 69.0 & 62.8 & 57.9 & 49.5 & 39.5 & 55.7 &  56.3  \\ 
&LookupViT$_{5\times5}$        & 70.6 & 64.2 & 58.6 & 49.7 & 38.8 & 56.4 & 55.3  \\ 
&LookupViT$_{7\times7}$      & 72.0 & 65.8 & 60.7 & 52.0  & 41.6 & 58.4 & 52.8 \\ 
&LookupViT$_{10\times10}$      & 72.9 & 67.1 & 62.2 & 53.9  & 43.6 & \textbf{59.9} & \textbf{50.8} \\ 
\midrule
\parbox[t]{6mm}{\multirow{4}{*}{\small L/16}} 
&ViT \cite{dosovitskiy2020image} & 68.3 & 62.4 & 57.3 & 49.2 & 39.3 & 55.3 & 54.7 \\
&TokenLearner$_{3/4}$ \cite{ryoo2021tokenlearner} & 69.6 & 63.4 & 58.0 & 48.9 & 38.2 & 55.6 &  56.4  \\ 
&LookupViT$_{5\times5}$        & 71.4 & 65.5 & 60.6 & 52.3 & 42.3 & 58.4 & 52.9  \\ 
&LookupViT$_{7\times7}$      &  72.1 & 66.4 & 61.8 & 53.8 & 44.2 & 59.7 & 51.2\\ 
&LookupViT$_{10\times10}$      &  72.4 & 67.0 & 62.7 & 54.6 & 45.2 & \textbf{60.4} & \textbf{50.4}\\ 

\bottomrule
\end{tabular}
\end{table*}

In Figure 1b of the main text, we validate that the better performance of LookupViT in adversarial settings is not a mere artifact of its better discriminatory power, meaning that it has additional robustness properties compared to a vanilla ViT model. In order to show this, we analyse the deviation in the feature when the image is corrupted, for both vanilla ViT and LookupViT. Mathematically, for every image ($\boldsymbol{\mathrm{X}}$), we compute the normalized feature deviation \[\frac{||\boldsymbol{\mathrm{F}}(\boldsymbol{\mathrm{X}})-\boldsymbol{\mathrm{F}}(\boldsymbol{\mathrm{X}}_c)||_2}{||\boldsymbol{\mathrm{F}}(\boldsymbol{\mathrm{X}})||_2}\]
where $\boldsymbol{\mathrm{X}}_c$, $\boldsymbol{\mathrm{F}}$ are the corrupted image version, and the model respectively. $|| . ||_2$ denotes the $L_2$ norm. A lower value for this metric signifies greater robustness to perturbations. The distribution of feature deviation for LookupViT has a lower mean than vanilla ViT. Moreover, with increasing severity, the mean feature distance increases more for ViT than LookupViT.

\subsection{Performance Analysis on Something-Something-V2}
\label{app:ssv2}

In Table 4 of the main text, we demonstrate that on the Something-Something V2 dataset \cite{goyal2017something}, the performance improvements due to regularization when using the ViViT Factorised Encoder \cite{arnab2021vivit} model do not translate to the ViViT-Base model. In this section, we extensively try to enhance the vanilla ViViT-Base model. Table \ref{table:ssv2_supp} lists the performance of the ViViT-Base model, when employed with different initialisation and regularisation strategies. The model is initialised using either a  Kinetics 400 or a ImageNet21k pretrained checkpoint. We analyse these variants both in presence and absense of regularisation parameters (label smoothing, mixup, stochastic droplayer). While the norm is to initialise with a Kinetics 400 checkpoint using regularisation parameters  as mentioned in ViViT \cite{arnab2021vivit}, but experiments show that for the base model, the unregularised variant performs better. However, LookupViT based ViViT-Base models, with the standard set of parameters outperforms all of these. As we can see, even after all the improvements on vanilla ViViT model, LookupViT with half the FLOPs exhibits better performance.

While the Kinetics-400 dataset suffers from static bias \cite{li2019repair}, SSv2 does not and thus performance on this dataset is often used as a measure of being unbiased \cite{li2023mitigating}. Our method's better performance on SSv2 can be attributed to it being less biased.

\begin{table*}[hbtp!]
\centering
\caption{Comparison of LookupViT-B/16 based ViViT \cite{arnab2021vivit} model with fine-tuned ViViT-Base on the SomethingSomethingV2 \cite{goyal2017something} dataset.}
\label{table:ssv2_supp}
\scriptsize
\setlength{\tabcolsep}{8pt}
\resizebox{\textwidth}{!}{
\begin{tabular}{l cc | c | c c }
\toprule
\multirow{2}{*}{Method} & \multicolumn{2}{c}{SomethingSomethingV2} & \multirow{2}{*}{GFLOPs/view} & \multirow{2}{*}{K400 Init} & \multirow{2}{*}{Reg} \\
                        & 1 view & 12 views   \\ 
\midrule
\parbox[t]{6mm}{\multirow{4}{*}{\small ViViT \cite{dosovitskiy2020image}}}
& 50.8 &  52.8 & 455.1 & $\times$ & \checkmark \\
& 51.2 & 52.3 & 455.1 & \checkmark & \checkmark \\
& 54.5 & 56.5 & 455.1 & $\times$ & $\times$ \\
& 55.0 & 57.1 & 455.1 & \checkmark & $\times$ \\

LookupViT$_{9\times9\times8}$            & 56.3 & 58.3 & 226.5 & \checkmark & \checkmark \\ 
LookupViT$_{8\times8\times16}$        & 57.0 & 59.2 & 311.2& \checkmark & \checkmark\\ 
LookupViT$_{9\times9\times16}$      & 57.1 & 59.6 & 375.8& \checkmark & \checkmark \\ 
\bottomrule
\end{tabular}
}
\end{table*}

\subsection{Comparison with other efficient networks on ImageNet-1k}

While we compare our method against three key architectures - ViT, Token Learner and Perceiver, in the main paper, we further contrast the performance of LookupViT against some more techniques in Table \ref{table:additional}. Since some of these methods report results using different training frameworks (ViT/DeIT/DeIT3), we report relative gains in accuracy along with relative computational savings for a fair comparison.

\begin{table}[b!]
\centering
\caption{Comparisons with respective B/16 baselines on IN-1k}
\scriptsize
\label{table:additional}
\setlength{\tabcolsep}{5pt}
{
\begin{tabular}{lccccc}
\toprule
Model $\rightarrow$ & CViT \cite{you2023castling} & ST \cite{pan2023slide} & Compress \cite{wei2023joint} & CAP \cite{kuznedelev2024cap} & LViT-B/16$_{7\times7}$ \\
\midrule
Accuracy (\% $\uparrow$) &  +0.8 & +0.7 & -0.6 & -0.2 & \textbf{+1.1} \\
GFLOPS ($\downarrow$)   &  1x & 1x & 0.67x & 0.5x & \textbf{0.67x}\\
\bottomrule
\end{tabular}
}
\end{table}

\subsection{Few-shot Transfer Results}

In this section, we compare the generalization properties of LookupViT as compared to ViT, through few shot evaluations on standard image datasets like Birds, CalTech, CIFAR100, ImageNet-1k, and Pets. For this set of experiments, we use models pre-trained on ImageNet-21k and evaluate them on these datasets under three settings - 1-shot, 5-shot and 25-shot. The results are presented in Table \ref{table:fewshot}. It can be noted that our method outperforms the base model performance on a lot of these settings, often by significant margins.

\begin{table}[hbtp!]
\centering
\caption{Few-shot eval, ViT \& LViT (B/16), IN-21K pre-training (1s: 1 shot)}
\label{table:fewshot}
\scriptsize
\setlength{\tabcolsep}{5pt}
\begin{tabular}{l c c c | c  c c | c c c | c c c | c c c}
\toprule
  & \multicolumn{3}{c}{Birds} & \multicolumn{3}{c}{CalTech} & \multicolumn{3}{c}{CIFAR100}  & \multicolumn{3}{c}{IN-1K} & \multicolumn{3}{c}{Pets} \\
                        & 1s & 5s & 25s  & 1s & 5s & 25s  & 1s & 5s & 25s & 1s & 5s & 25s & 1s & 5s & 25s\\ 
\midrule
 ViT                     & 63 & \textbf{80} & \textbf{87} & 80 & 86 & 91 & 52 & 77 & 84 & 52 & 71 & 77 & 70 & 86 & 92\\
 LViT{$_{10\times10}$}     & \textbf{65} & 79 & 85 & \textbf{83} & \textbf{87} & \textbf{91} & \textbf{61} & \textbf{81} & \textbf{86} & \textbf{64} & \textbf{77} & \textbf{80} & \textbf{76} & \textbf{87} & \textbf{92}\\
\bottomrule
\end{tabular}
\end{table}

\subsection{Attention Maps across Image Sizes and Primary Token Count}
\label{app:im1k_attn}

\begin{figure}[b!]
\includegraphics[width=\textwidth]{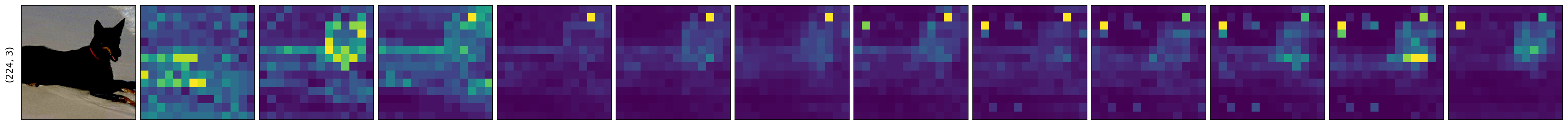}\\
\includegraphics[width=\textwidth]{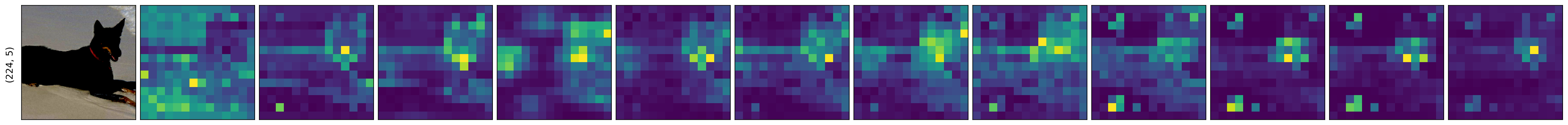}\\
\includegraphics[width=\textwidth]{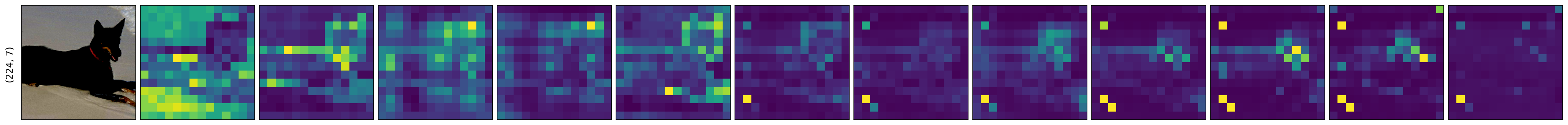}\\
\includegraphics[width=\textwidth]{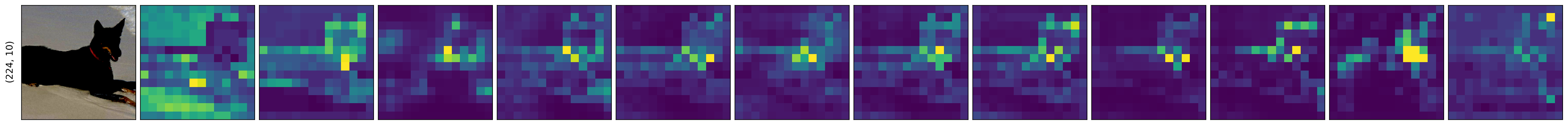}\\
\includegraphics[width=\textwidth]{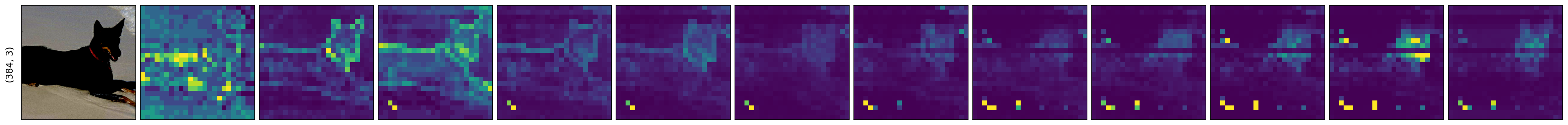}\\
\includegraphics[width=\textwidth]{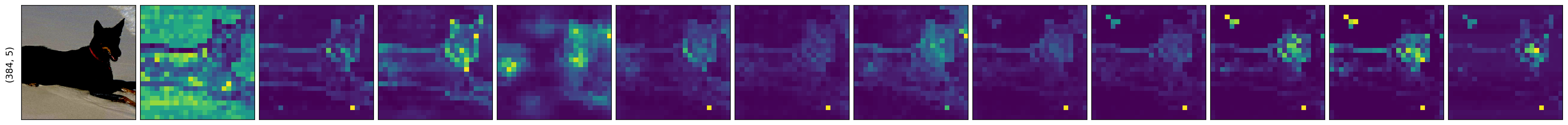}\\
\includegraphics[width=\textwidth]{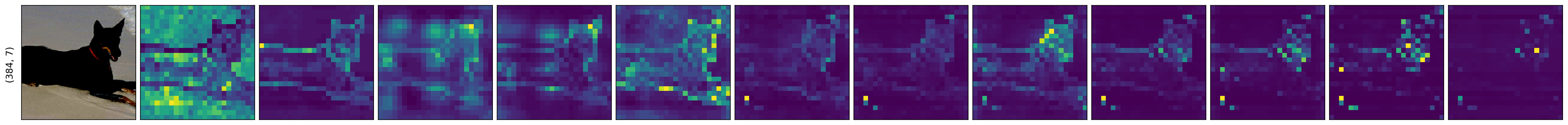}\\
\includegraphics[width=\textwidth]{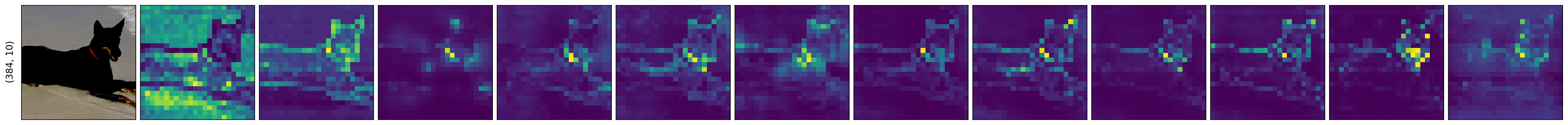}\\
\includegraphics[width=\textwidth]{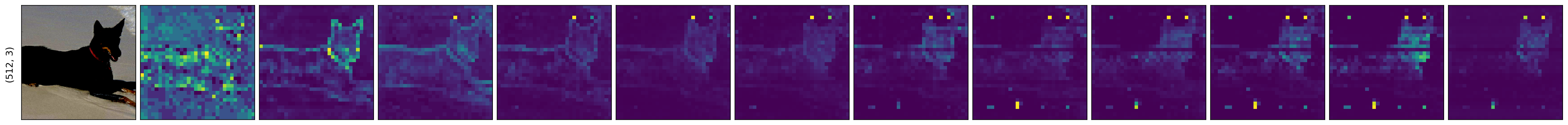}\\
\includegraphics[width=\textwidth]{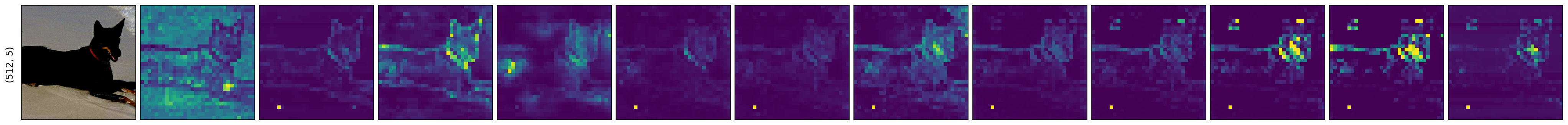}\\
\includegraphics[width=\textwidth]{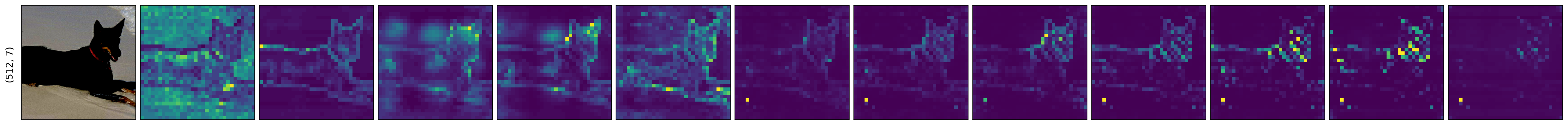}\\
\includegraphics[width=\textwidth]{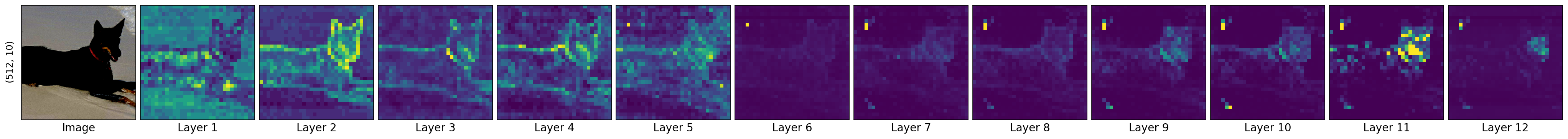}

\caption{The above set of figures depict the layerwise attention maps for various image resolutions and the number of primary tokens. The label on the left (R, S) indicates a image resolution "R" and number of primary tokens "S" along each of the 2D axis. Note that the attention maps become finer as the image resolution goes up.}
\label{fig:attwres}
\end{figure}

Figure \ref{fig:attwres} depicts the attention maps computed by the LookupViT-B/16 model trained on ImageNet-1k for different image resolutions and number of primary tokens. Each row is annotated with the corresponding values for these two parameters on the left. Each row represents the image, followed by the layerwise attention maps, averaged over all the attention heads, as well as over the primary tokens.

As the image resolution goes up, the cross-attention maps become finer in the sense that their representation power goes up. This is consistent with vanilla ViT models. However, the number of primary tokens being another choice to be made, there are two things at play in LookupViT. With a constant primary token count and increasing image resolution, the down-sampling ratio goes up and thus the information bottleneck becomes more stringent. A weak signal of the argument can be seen in Figure \ref{fig:attwres}, where the attention maps for (384, 3) look ``stronger" than those for (512, 3). However, the increasing accuracy trend with resolution for all patch sizes, as seen in Figure 1b (of paper), indicates that this effect is well subdued. 

Another interesting detail to note here is the identification of salient objects in the early layers itself. This allows the later layers to concentrate on the relevant regions. Analogous to ViT, information is repurposed across tokens in the later layers for easier internal computation, which may not be otherwise intuitive or aligned with the image \cite{darcet2023vision}. This partially explains the artifacts in the attention maps, and works from literature \cite{darcet2023vision} can mitigate them for better visualization. 

\subsection{Attention Maps on Something-Something-V2 Video Classification}
\label{app:ssv2_attn}

\begin{figure}[htbp!]
\includegraphics[width=\textwidth]{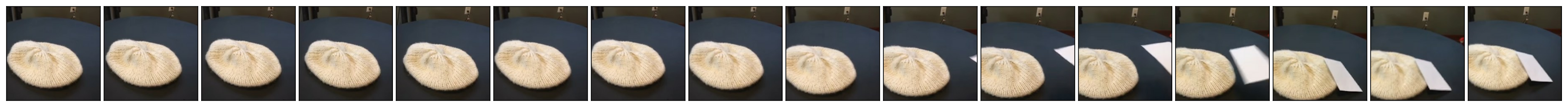}\\
\includegraphics[width=\textwidth]{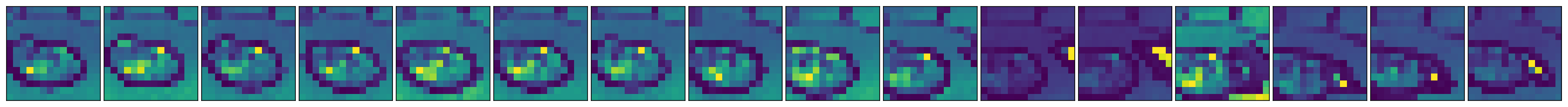}\\
\includegraphics[width=\textwidth]{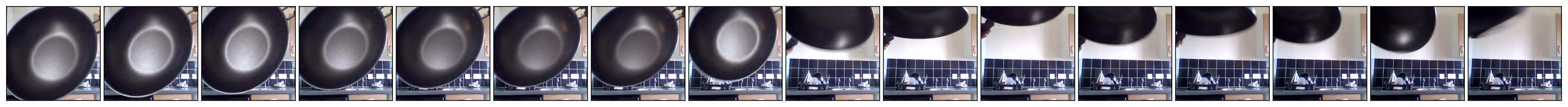}\\
\includegraphics[width=\textwidth]{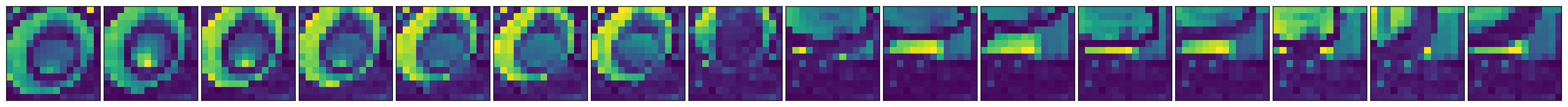}\\
\includegraphics[width=\textwidth]{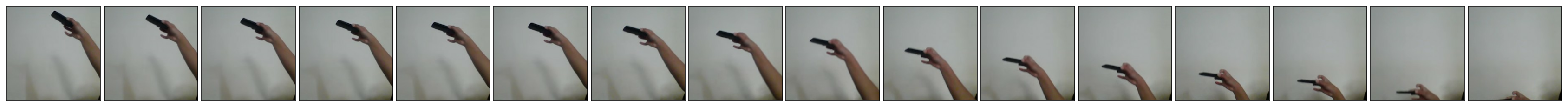}\\
\includegraphics[width=\textwidth]{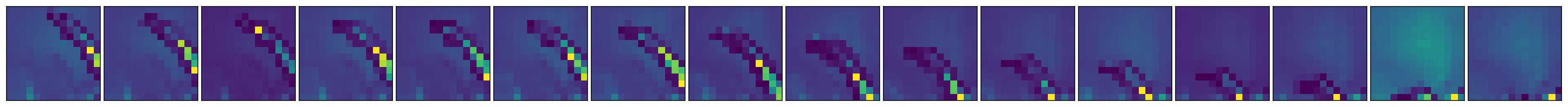}

\caption{Attention maps computed by LookupViT model during video classification on the Something-Something-v2 dataset. Each pair of rows represent the changes in video frames and corresponding attention maps with time (x-axis). The attention maps are taken from the first layer of the model and are averaged along the heads and the primary tokens.}
\label{fig:vidatt}
\end{figure}

Figure \ref{fig:vidatt} depicts the attention maps computed by LookupViT for some of the video inputs from the Something-Something-v2 dataset. In the case of images, the attention maps from the second layer of the model best correlated with the image features. However, in the case of videos, we observe that the first layer best represents the local information. In Figure \ref{fig:vidatt}, the sampled video frames and the attention maps at the same temporal stamps have been presented as sequences of images. 

The first example supports the model's capability to readjust its focus to suddenly occuring motion. Towards the end of this video, a piece of paper falls into view and the attention maps quickly adjust and focus on the falling piece of paper. The second example is a good representation of the model's capability of identifying and focusing on the salient object. This is evident by the fact that the attention maps neglect the static but complicated background effectively and only focus on the moving object (cook pan) in the foreground. The third example represents the model's capability to identify small objects. Even with a very coarse attention map, the moving hand in the video is effectively traced by the attention values. These observations provide a visual demonstration of the model's capabilities and support its applicability in a variety of scenarios.

\end{document}